\def\year{2021}\relax
\documentclass[letterpaper]{article} 
\usepackage{aaai21}  
\usepackage{times}  
\usepackage{helvet} 
\usepackage{courier}  
\usepackage[hyphens]{url}  
\usepackage{graphicx} 
\usepackage{natbib}  
\usepackage{caption} 
\usepackage{microtype}
\usepackage{subfigure}
\usepackage{booktabs} 
\usepackage{amsmath}
\usepackage{amsfonts}
\usepackage{bm}
\usepackage{siunitx}
\usepackage{makecell}
\usepackage[inline]{enumitem}
\usepackage{todonotes}
\newcommand{\ra}[1]{\renewcommand{\arraystretch}{#1}}

\title{A Reinforcement Learning Environment For Job-Shop Scheduling}
\author {
    Pierre Tassel, \textsuperscript{\rm 1} 
    Martin Gebser, \textsuperscript{\rm 1} \textsuperscript{\rm 2} 
    Konstantin Schekotihin \textsuperscript{\rm 1} \\
}
\affiliations {
    \textsuperscript{\rm 1} Department of Artificial Intelligence and Cybersecurity\\
    University of Klagenfurt, Klagenfurt, Austria \\
    \textsuperscript{\rm 2} Institute of Software Technology\\
    Graz University of Technology, Graz, Austria \\
    pierre.tassel@aau.at, martin.gebser@aau.at, konstantin.schekotihin@aau.at
}
\begin{document}

\maketitle
\begin{abstract}
Scheduling is a fundamental task occurring in various automated systems applications, e.g., optimal schedules for machines on a job shop allow for a reduction of production costs and waste.
Nevertheless, finding such schedules is often intractable and cannot be achieved by Combinatorial Optimization Problem (COP) methods within a given time limit. 
Recent advances of Deep Reinforcement Learning (DRL) in learning complex behavior enable new COP application possibilities. 
This paper presents an efficient DRL environment for Job-Shop Scheduling -- an important problem in the field.
Furthermore, we design a meaningful and compact state representation as well as a novel, simple dense reward function, closely related to the sparse make-span minimization criteria used by COP methods.
We demonstrate that our approach significantly outperforms existing DRL methods on classic benchmark instances, coming close to state-of-the-art COP approaches.
\end{abstract}

\section{Introduction}
\label{introduction}

Deep Reinforcement Learning (DRL) has gained an increasing interest in the AI community, thanks to many recent successes like Atari \cite{atari}, AlphaGo \cite{Silver2017}, or AlphaStar \cite{Vinyals2019}.
Most of the approaches in this field focus on well-established benchmarks from the OpenAI gym library or video games (Atari, Starcraft, Dota, etc.). 
Given these successful applications, researchers soon recognized the potential benefits of DRL in practical domains, such as  Combinatorial Optimization Problems (COPs) \cite{BENGIO2021405}. 

\begin{figure}[t]
\begin{center}
\includegraphics[width=\linewidth]{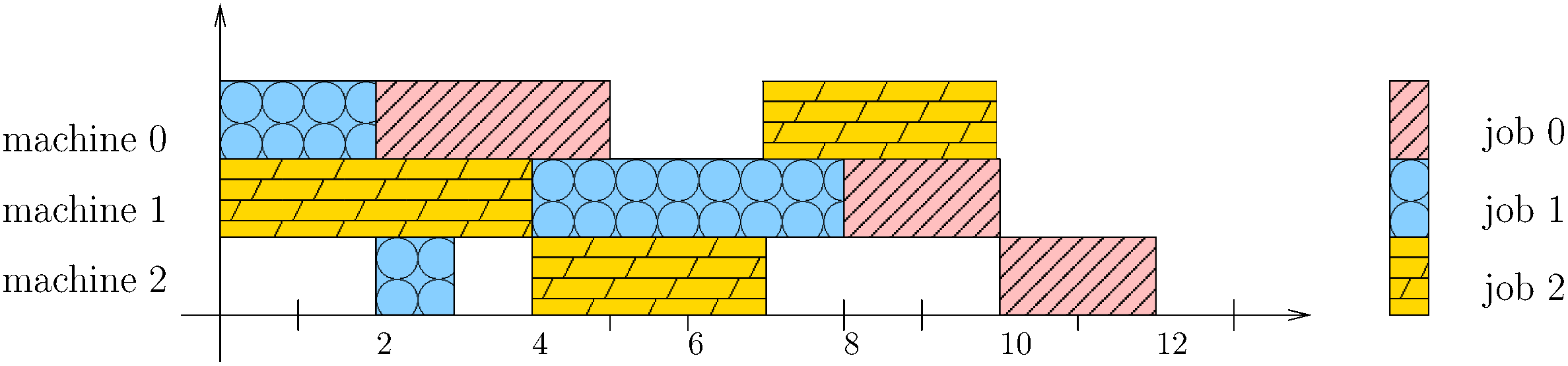}
\caption{Example solution for a small instance composed of three jobs and three machines. Each machine sees each job once, and the time spent on each operation varies from one job to another.}
\label{Fig:small_solution}
\end{center}
\end{figure}

In this paper, we present a DRL-based approach to Job-Shop Scheduling (JSS), an important problem that was among the first COPs ever studied \cite{eco1996}. Its application domains are wide, from the ordering of computation tasks to the scheduling of manufacturing systems.
In its classic variant, each instance of JSS comprises two sets of constants representing jobs $\mathcal{J}$ and machines $\mathcal{M}$.
Each job $J_i \in \mathcal{J}$ must go through each machine in $\mathcal{M}$ in a specific order denoted as $O_{i1}\rightarrow\dots\rightarrow O_{in_i}$, where each element $O_{ij}$ ($1\leq j \leq n_i$) is called an operation of $J_i$ with a processing time $p_{ij} \in \mathbb{N}$. The binary relation $\rightarrow$ is called the precedence constraint.
There is a total of $|\mathcal{J}|\times|\mathcal{M}|$ operations to perform, and their executions are non-preemptive. Also, no machine can perform more than one operation simultaneously.
Fig.~\ref{Fig:small_solution} represents a solution for a small instance composed of three jobs and three machines.

The application of Reinforcement Learning (RL) to JSS provides several advantages.
First, it is more flexible than traditional priority dispatching rule heuristics, whose performance can vary significantly from instance to instance. In addition, the development of such heuristics is quite tedious since it requires a lot of specialized knowledge about a scheduling instance to be effective.
Unlike classic COP methods, such as linear programming (LP) or constraint programming (CP), RL environments can model stochastic decisions, e.g., random factory outage, non-deterministic job re-entry, random processing time, etc., which simulate the conditions faced by real scheduling systems.
Second, in contrast to traditional scheduling methods that focus on the given set of jobs only, RL provides a possibility to incrementally schedule the incoming jobs as they appear in the queue by considering the impact of a schedule for known jobs on the new ones. 
Finally, the most promising possibility offered by RL is the concept of lifelong learning \cite{chen2018lifelong}, where an agent will not only learn to optimize one specific JSS instance but reuse what it has learned from previous instances. 
This property is essential for industrial problems where instances usually share a lot of similarities. 

In this paper we make the following contributions:
\begin{itemize}
\item We propose to model JSS as a single-agent RL problem, where the agent is a dispatcher that needs to choose which job to work on at each step.
In particular, we use the actor-critic Proximal Policy Optimization (PPO) algorithm \cite{Schulman2017Ppo} to learn a policy, i.e., a function that maps a state to an action probability distribution, and approximate the state-value function, i.e., the excepted cumulative reward given a state.
\item We design an environment with a meaningful and compact state representation, as well as a novel and simple dense reward function, closely related to the sparse make-span minimization objective of COP methods.
\item Experiments on instances that are hard for COP methods under time constraints indicate that the application of RL algorithms in our environment results in solutions close to those of the best scheduling techniques on the market and far better than results reported in the RL literature before.
\end{itemize}

\section{Background}
\label{Background}

In this section, we provide some mathematical background for Markov Decision Processes (MDPs) and RL, give details of the PPO algorithm, and discuss other RL scheduling approaches found in the literature.

\paragraph{Markov Decision Processes}

MDPs are discrete-time stochastic control processes used to model decision making.
In this process, the outcome is partly random under the control of a decision-maker.
Formally, an MDP is defined as a 4-tuple $M=(S,A,P_{a},R_{a})$ where $S$ is a set of states (state space), $A$ is a set of actions (action space), $P_{a}(s, s')$ the probability to go from state $s$ to $s'$ by taking action $a$ (in our case the environment is deterministic, so the probability is set to 1) and $R_{a}(s, s')$ the reward we get after transitioning from state $s$ to $s'$.

\paragraph{Reinforcement Learning}

The goal of an RL training algorithm is to learn a policy $\pi$ that maps each state to an action with the intent to maximize the expected cumulative reward. Given a policy $\pi$, we can compute the value $V^{\pi}$ of this policy for a state $s_0$ as:
\begin{align}\nonumber
V^\pi(s_0)&=\mathbb{E}^\pi\left[\sum_{t=0}^{+\infty}\gamma^t R(s_t,a_t)\right]
\end{align}
with $\mathbb{E}^\pi$ the expectation over the distribution of the admissible trajectories $(s_0,a_0, r_0, s_1, a_1, r_1, \dots)$ obtained by sampling the actions from the policy $\pi$ and $\gamma$, the discount-rate, a hyper-parameter that controls how far the agent looks into the future. A policy is optimal if no other policy yields a higher return value.

A value-based algorithm tries to approximate the \emph{state-value function}. When one has learned this function, the policy can be defined as selecting the action leading to the state with the maximum value \cite{atari}.
In contrast, a \emph{policy-based algorithm} tries to directly learn the policy maximizing the cumulative reward by increasing the probability of actions that yield a good return \cite{sutton2018reinforcement}.
State-of-the-art policy-based algorithms combine these two approaches by learning the policy and the state-value function. These algorithms are called \emph{actor-critic methods} \cite{Mnih2016}.

\subsection{Proximal Policy Optimization}

Policy-gradient algorithms learn the policy by computing an estimator of the policy gradient through experience sampling and plugging it into a Stochastic Gradient Ascent (SGA) algorithm. The gradient estimator is of the form:
\begin{align}\nonumber
    \nabla_{\theta} J(\pi_{\theta}) = 
    \displaystyle \mathop{\mathbb{E}}_{\tau \sim \pi_{\theta}} \left[{\sum_{t=0}^{T} \nabla_{\theta} \log \pi_{\theta}(a_t|s_t) A^{\pi_{\theta}}(s_t,a_t)}\right]
\end{align}
where $\pi_{\theta}$ is a stochastic policy, and $A^{\pi_{\theta}}$ is an estimator of the advantage function at time-step~$t$.
Vanilla Policy-Gradient algorithm REINFORCE \cite{Williams1992} uses a complete roll-out as an unbiased estimator, but this estimator suffers from high variance. Actor-Critic methods overcome this by training a neural network to estimate the state-value of the $n^\text{th}$ state in the future. It avoids doing roll-out and trades off variance for bias \cite{Mnih2016}.

A fundamental property of Policy-Gradient algorithms is that update steps computed at any specific policy $\pi_{\theta}$ only guarantee predictiveness in a neighborhood around~$\theta$.
Vanilla Actor-Critic methods tend to suffer from ``policy-crash" (i.e., a performance collapse) if the step size is inadequate. Policy-Gradient algorithms keep policy updates close in the parameter space. But even seemingly small differences in the parameter space can have a substantial impact and 
deteriorate the policy performance \cite{Schulman2015,PPOOpenAI}.
PPO tries to take the biggest possible improvement step without stepping so far that we accidentally cause performance collapse. It does so by taking multiple steps of SGA on a clipped objective function. Here the loss function $L$ is given by:
\begin{align}\nonumber
& L = \min\left(
\frac{\pi_{\theta}(a|s)}{\pi_{\theta_k}(a|s)}  A^{\pi_{\theta_k}}(s,a), \;\;
g(\epsilon, A^{\pi_{\theta_k}}(s,a))
\right)
\\\nonumber
& g(\epsilon, A) = \left\{
    \begin{array}{ll}
    (1 + \epsilon) A & A \geq 0 \\
    (1 - \epsilon) A & A < 0
    \end{array}
    \right.
\end{align}
in which $\theta$ and $\theta_k$ are the parameters of the new and the old policy, respectively, and $\epsilon$ a (small) hyper-parameter which roughly says how far away the new policy is allowed to go from the old one.

\subsection{Related Work}

Recently, DRL has gained attraction as an end-to-end approach to solve COPs. From the Traveling Salesman Problem 
over Graph Optimization to the Satisfiability problem, 
the number of DRL applications is increasing sharply. However, applications of DRL to scheduling problems are more recent and limited.

One classic approach to model scheduling problems is to consider each machine as an agent. 
\citet{Waschneck2018} model JSS by a multi-agent system, 
where agents standing for machines are trained one after the other using Deep Q-Network (DQN) learning. An evaluation of their approach on a non-public instance set for a semiconductor factory shows that it is resilient to disruptions. It cannot beat heuristics but reaches expert-level performance after two days of training.
Another multi-agent approach was suggested in \cite{Liu2020}, where the authors applied Deep Deterministic Policy Gradient to train the agents. 
On small to mid-size instances---maximum of \num{20} jobs and \num{15} machines with known optimal solutions---their approach performs better  than dispatching rules, even though First-In-First-Out (FIFO) comes very close. Similar to \citet{Waschneck2018}, \citet{Liu2020} focus on flexibility and propose a scenario with random machine breakdowns.

While the multi-agent approach seems appealing, it can be hard to learn a policy where agents collaborate and share 
a global vision on the schedule to find good solutions.
The disjunctive graph representation of JSS is a modeling approach allowing for such a global vision.  \citet{zhang2020learning} model the problem by a single agent and exploit the disjunctive graph of a JSS instance to design an embedding of states using a Graph Neural Network. 
Moreover, the authors use a Multi-Layer Perceptron (MLP) to compute a scalar score for each action based on the graph embedding and apply a softmax function 
to output an action probability distribution. The advantage of this approach is that the policy network is independent of the size of an instance. It can be trained on small instances and then generalize to bigger ones.
Similarly, \citet{9218934} use the disjunctive graph to encode states as an image of size $|\mathcal{J}|\times|\mathcal{M}|$ representing the total number of operations in the instance. Channels of the image represent three features: processing time, the schedule at the current time-step, and machine utilization. Such an image is passed to a convolutional neural network that acts as a state-action value function approximator. The action space is defined as a set of \num{18} different dispatching rules that the agent can select. The reward function, like ours, reflects the impact of the job allocation on machine utilization. However, the disjunctive graph formulation makes computing a reward function, for which \citet{9218934} compare the average machine utilization before and after taking action,
harder than our representation. 
Finally, \citet{dqnScheduling} use a multi-class DQN method to solve JSS instances. Their agent learns to assign one of \num{7} dispatching rules for each machine. 
Their evaluation results on small to mid-size instances---maximum of \num{20} jobs and \num{20} machines with known optimal solutions---show that, although the suggested approach found better solutions than the individual dispatching rules, they are still far from being optimal.

\section{Job-Shop Scheduling Environment}
\label{jss_env}

In this section, we first introduce the basics of our JSS environment and explain its internal working. 
Then we detail our reward function and define the state space. Finally, we present some implementation details of our environment. 

\subsection{Basic Environment} 

We consider JSS as a single-agent problem, where a dispatcher chooses a job to work on at each time-step.
We define the action space as the discrete set of jobs plus an extra action, called No-Op, to go to the next time-step without scheduling any operation:
$A = \{J_{0}, ..., J_{|\mathcal{J}| - 1}, \text{No-Op}\}$. 

In some cases, we cannot allocate a job at each step since some jobs may already be allocated to a machine, the machine can be in use, or we have completed all the job's operations. To comply with these constraints, our environment also uses a Boolean vector indicating which actions are legal.
Each time the agent allocates a job, it adds the assigned job to the ordered stack of the next time-step we will visit. When we cannot allocate any further job or decide for No-Op at a certain time-step, we iteratively go to the next time-step in order to allocate jobs again (i.e., a machine becomes free).

\begin{figure}[tbh]
\begin{center}
\includegraphics[width=\linewidth]{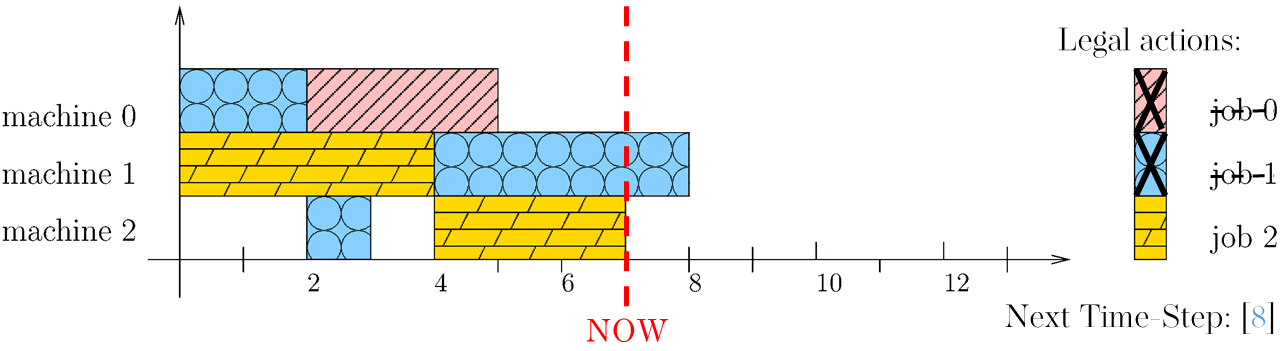}
\caption{Representation of the environment's internal state comprising the information about the current time-step, the allocation of jobs to machines, the jobs that can be allocated now and the future time-steps.}
\label{Fig:env_rep}
\end{center}
\end{figure}

In Fig.~\ref{Fig:env_rep}, we illustrate the internal status of the environment. The vertical red line represents the current time-step. We can allocate job $J_2$ (in yellow) to machine $M_0$ (the machine we will allocate the job to is not given explicitly by the environment). Or, as we have only one legal action, we can go to the next time-step $T_8$ for machine $M_1$ to become free and allocate job $J_0$ to it.

\subsection{Search-space Reduction}\label{par:non_final}

\begin{figure}[t]
\begin{center}
\includegraphics[width=\linewidth]{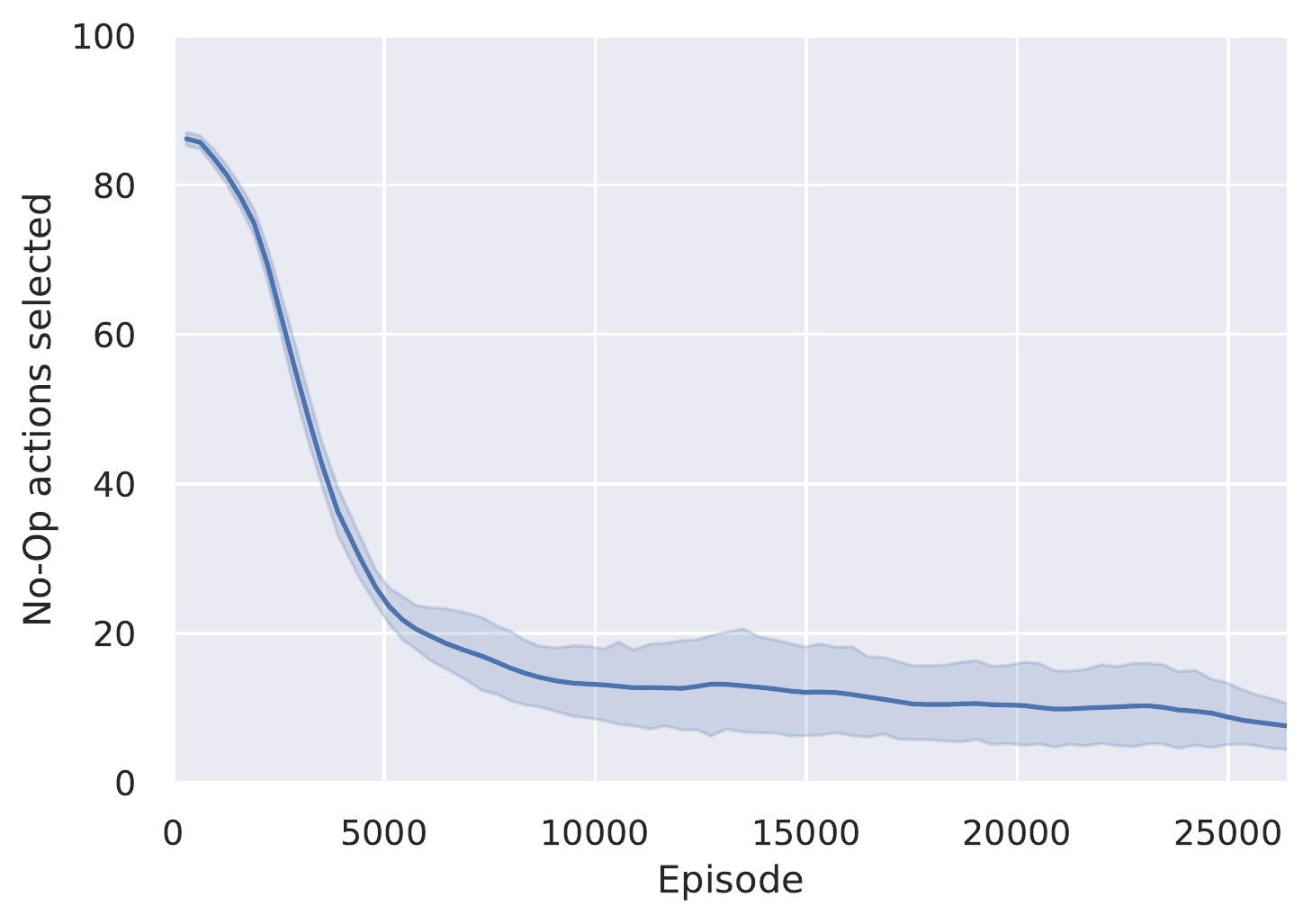}
\caption{The agent learns to use the No-Op action sparingly without avoiding it completely. The number of No-Op actions selected was measured using a running average over the last 2000 episodes for each of Taillard's instances. This plot represents the mean and 95\% confidence interval.}
\label{Fig:no_op_selection}
\end{center}
\end{figure}

To achieve good performance in a short time, we need to guide our agent to avoid exploring sub-optimal solutions.

Considering two jobs, $J_1$ and $J_2$, to allocate to the same machine, if $J_1$ is at its final operation (i.e., the job has already completed $|\mathcal{M}|-1$ operations and needs only one more operation to be finished) and $J_2$ is not, it is always better to allocate $J_2$, as it will still need another machine afterwards. We refer to this optimization as \textbf{non-final prioritization}.

Compared to other actions, the No-Op action deserves particular attention. This action is complex to learn, as it often leads to worse results than greedy policies (disregarding No-Op and allocating jobs to machines whenever possible) if not used with care. If we do not guide the agent to learn how to use No-Op, it will stop using it and become a greedy agent because the weight of the No-Op's logit will tend to~\num{0}.

The first rule we apply marks jobs that could be allocated to a machine as illegal when No-Op is used instead, as long as we do not allocate a new job to the machine (i.e., a job that was not legal before the No-Op). We do this because it is always better (or equivalent) to allocate a job $J$ directly at time-step $T$ rather than waiting for $X > 0$ time-units and then allocating $J$ at time-step $T + X$. 
This restricts the use of the No-Op action, as we insist on the availability of
some new job to allocate in the future, or the considered machine will be blocked indefinitely otherwise.

To restrict No-Op further, for each machine, we compute the minimum duration $D$ of each legal job we can schedule. Next, we check whether we will have a new job to allocate to the machine in less than $D$ time-units. This rule is based on the observation that, if we make a pause longer than $D$, it would have been better to allocate a job of duration $D$ before. 
Additionally, the new job we need to allocate next should not be at its final operation because waiting for a job rejected by non-final prioritization (see above) is pointless.

To prevent almost certainly bad uses of No-Op and avoid time-consuming No-Op checks, we disallow the No-Op action if there are
\begin{enumerate*}[label=\emph{(\arabic*)}]
\item four or more machines with some allocatable job, or
\item five or more allocatable jobs (more than \si{15\percent} of the jobs/machines available in total).
\end{enumerate*}
If the agent still takes the No-Op action, we iterate over the next time-steps until some new job becomes allocatable.
In Fig.~\ref{Fig:no_op_selection}, we see that the agent learns the potentially deleterious effect of the No-Op action on the solution quality quickly and uses it sparingly towards the end of the training episodes.

\subsection{Problem Symmetries}

JSS comprises a lot of symmetries that can increase the search space unnecessarily. Our environment considers two of them:
\begin{enumerate*}[label=\emph{(\arabic*)}]
    \item interchangeability of allocations at a time-step, and 
    \item the execution of jobs and the No-Op action on a machine.
\end{enumerate*}
In the first case, e.g., an agent might allocate a job $J_1$ to machine $M_1$, and then job $J_2$ to machine $M_2$ at the same time-step. This allocation order is symmetric to allocating $J_2$ to $M_2$ first and then $J_1$ to $M_1$. 
Such symmetry is less pronounced in our environment because the applicability of the No-Op action depends on the number of jobs and machines available, so that the order of allocation can matter.
Note that we cannot break this symmetry by ordering the machines and first allocate some job to the machine with the smallest identifier. Such a model leads to a multi-agent system with policy sharing, thus losing the global view and 
specializing the No-Op action to each individual machine.

The No-Op action itself introduces some symmetry. If a job is allocated to a machine that needs to wait afterwards, we can either complete the job and wait, or first wait and then allocate the job. As explained in Section~\ref{par:non_final}, we break such symmetry by marking 
allocatable jobs as (temporarily) illegal when No-Op is used, which forces either immediate
job execution or future allocation of some new job instead.

In addition, there may be intra-machine symmetries, when the execution of two non-final operations of different jobs can be swapped due to waiting times before allocating their respective successor operations.
Such dynamic symmetries cannot easily be broken at the time-step of job allocation.
Non-final prioritization (see Section~\ref{par:non_final}) breaks these symmetries partially, whenever it is equivalent to allocate a final operation first and a non-final job afterwards.

\subsection{Reward Function}

The traditional metric to evaluate a solution is the maximum make-span across all the jobs, i.e., the time needed to have finished every job. However, this metric might lead to a sparse reward function, which provides feedback only at the end of the episode. As a result, the agent might struggle to determine the impact of each action it has taken on the global outcome.

To improve learning, we have designed a dense reward function based on the \emph{scheduled area}. After each action, we compute the difference between the duration of the allocated  operations and the introduced holes---the idle time of a machine:

\begin{align*}
R(s, a) = p_{aj} - \sum_{m \in \mathcal{M}} \mathrm{empty}_{m}(s,s')
\end{align*}

where $s$ and $s'$ the current and next state respectively; $a$ the $j^\text{th}$ operation of $J_a$ with a processing time $p_{aj}$ scheduled (i.e., the action); $s'$ the next state resulting from applying action $a$ to state $s$; $empty_{m}(s, s')$ a function returning the amount of time a machine $m$ is IDLE while transitioning from state $s$ to $s'$.
Note, a re-scaled reward---$R(s, a)$ divided by the maximum operation length, turned out to be more suitable for training of neural networks in our experiments.

\begin{figure}[t]
\begin{center}
\includegraphics[width=\linewidth]{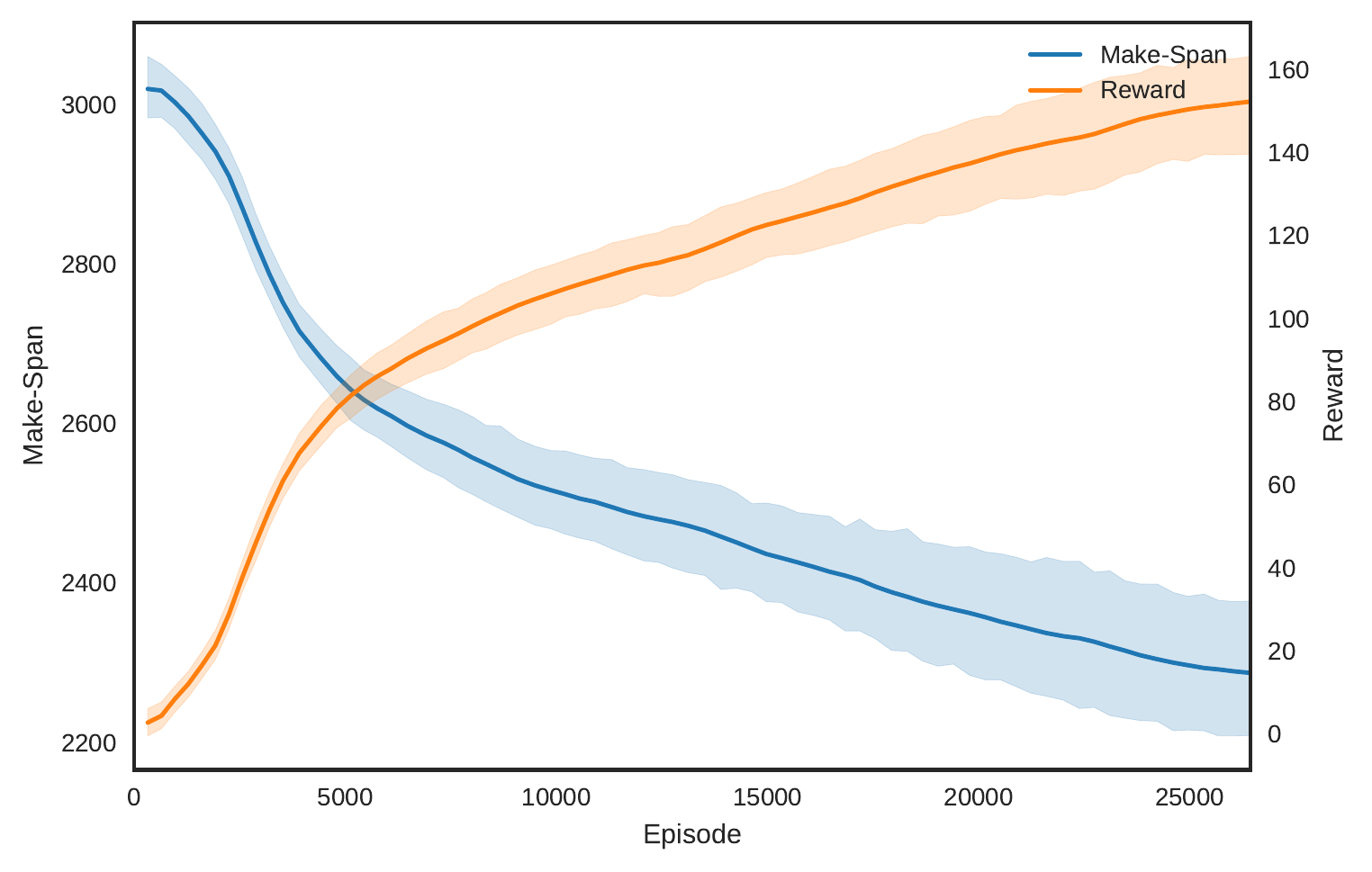}
\caption{There is a clear correlation between the real, sparse objective function (make-span) and our dense reward function. Both metrics were collected on Taillard's instances using a running average over the last 2000 episodes. The lines represent the means over all instances, and surroundings give the 95\% confidence interval.}
\label{Fig:reward_vs_makespan}
\end{center}
\end{figure}

For example, in Fig.~\ref{Fig:env_rep}, suppose we allocate job $J_2$ on machine $M_0$. In that case, we get as a positive reward---the length of the operation we have scheduled. Since there is no more available action, the environment will automatically jump to the next time-step. Nevertheless, this action creates a hole on machine $M_2$. As we currently are at time-step $T_7$, and we jump to time-step $T_8$, the negative reward will be \num{1}. We then sum the positive and negative rewards up to get the total reward.
Therefore, the reward function results in the minimization of the schedule area on a Gantt chart, i.e., minimize the holes for each machine and maximize machine usage. This is different from minimizing the make-span, but both criteria are closely related---a good solution has as few holes as possible.
This relation can be observed in Fig.~\ref{Fig:reward_vs_makespan} where we compare the evolution during the training of the mean make-span of a solution found and its cumulative reward. The figure represents an inverse variation relationship: the make-span decreases whenever the reward increases. This experiment provides sufficient evidence for the relation of these two criteria.

Nevertheless, this reward function is not perfect because there is still a delay between an action and its impact. For example, if we can allocate multiple jobs at a certain time-step, then only the last action of this time-step will create holes. As a result, only this action will get negative feedback. We mitigate this issue by using the algorithm that does multiples steps to compute each action's return. Hence, small delays in the reward will not affect it much as the number of actions at a time-step is relatively small. Another problem is that some choices done in the past can have a huge impact on the future outlook, e.g., force us to create a lot of holes. Unfortunately, this issue is inherent to the complexity of this problem.

\subsection{State Representation}

Our state representation is a $(\mathcal{J}\times7)$ matrix, containing, for every row (i.e. every job) 7 attributes:
\begin{enumerate*}[label=${(a_\arabic*)}$]
    \item A Boolean to represent if the job can be allocated.
    \item The left-over time for the currently performed operation on the job. This value is scaled by the longest operation in the schedule to be in the range $[0, 1]$.
    \item The percentage of operations finished for a job.
    \item The left-over time until total completion of the job, scaled by the longest job total completion time to be in the range $[0, 1]$.
    \item The required time until the machine needed to perform the next job's operation is free, scaled by the longest duration of an operation to be in the range $[0, 1]$. 
    \item The IDLE time since last job's performed operation, scaled by the sum of durations of all operations to be in the range $[0, 1)$.
    \item The cumulative job's IDLE time in the schedule, scaled as $a_6$ to be in the range $[0, 1)$.
\end{enumerate*}
This information is sufficient to reconstruct the current state of the schedule, and so, it respects the Markov condition. However, the reconstructed solution may not be unique but will be equivalent, i.e., leads to the same solution.
We have considered other attributes, such as the number of other jobs that need the same next machine and more global information about the process like the number of next time-steps, the number of legal actions, etc.
However, the number of attributes needs to stay short to avoid too much state computation time, i.e., the environment became slow, and avoid too much neural network computations, i.e., the agent became slow.
After experimentation, we have select these seven attributes out of other possible attributes. 

One attractive property of this representation is how it easily allows constructing dispatching rules. For example, the Most Work Remaining (MWKR) is equivalent to take the job with the largest value of the  $a_4$ attribute, i.e., the job that has the most left-over time until completion. Similarly, the First In First Out (FIFO) amounts to take the biggest value of the $a_6$ attribute, i.e., the job which was idle for the most time since its last operation.

\subsection{Problem Specification}

An episode is composed of a lot of steps, i.e., $\mathcal{O}(|\mathcal{J}|\times|\mathcal{M}|)$ and because of the No-Op operation, this is a lower bound. As the choice at the first steps can significantly impact the solution's outlook, the agent needs to use the experience it has learned during previous iterations to improve the next one.
More formally, the current state contains a lot of information from the previous step, and one cannot recover from a bad state.

We have applied some optimizations to make the environment fast enough. On one hand, we tried to keep the state representation as small as possible. On the other, various code improvements were made to speed up the step computation. Thus, we keep the state representation in memory and update only the attributes that need to be updated at each step rather than recomputing them.

\section{Method}
\label{method}

In this section we provide details of our approach and describe its implementation.

\subsection{Action Selection}
\label{action_selection}

Our environment gives us a tabular state represented by a matrix. To compute the action distribution and the state-value estimation, we flatten this matrix into a vector and pass it to the agent MLPs.

As explained previously, our environment has an additional specificity compared to a pure MDP model because we also get a mask of legal actions our agent can perform.
Different techniques can be applied to overcome this. One is to give a negative reward if the agent takes one illegal action and hopes it will learn to recognize them. This approach yields poor performance because it makes the problem even harder, as the agent has to learn at the same time to differentiate between a legal/illegal action and a good/bad action.

Our approach is to apply a mask on the output of a neural network to transform values of illegal action into a small negative number, i.e., the smallest representable number. It makes the illegal action probability very close to 0 when we apply the softmax function. This technique has already been studied in previous work and seems to yield the best results \cite{Huang2020}.

\subsection{Training Process}
\label{training}

Usually, the goal of reinforcement learning is to train an agent that can perform a specific task. For our setting, the problem is different, we still want to learn to schedule the production, but we are more interested in the agent's best solution.
To comply with this goal, we keep track of each solution's make-span found by the agent at the end of the episode.
Often in reinforcement learning, we train for a fixed number of steps or episodes, but, in our case, we have a time constraint. We have limited the training time to 10 minutes to have a realistic approach close to industrial constraints.

\subsection{Implementation}
\label{implementation}

We have implemented our approach using RLLib \cite{liang2018rllib} and Tensorflow \cite{tensorflow2015-whitepaper}. The environment is implemented with OpenAI Gym toolkit.\footnote{The code is available at
\url{https://github.com/prosysscience/JSSEnv}
}

We have used WandB \cite{wandb} Bayesian optimization to perform the hyper-parameter search and log all results provided in this paper.\footnote{\url{https://wandb.ai/ingambe/PPOJss/sweeps}}
Unfortunately, due to computing resources restriction, we have used only one instance (\textbf{ta41}) to perform our hyper-parameter search and used the best configuration on this unique instance for the whole group.
We have not performed extra hyper-parameter searches for Demirkol's instances and use the same ones we have used for Taillard's.

\subsection{Model and Configuration}
\label{model_configuration}

The action selection network $MLP_{\theta_\pi}$ and state-value prediction network $MLP_{\theta_v}$ do not share any layers. Both networks have \num{2} hidden layers with hidden dimension \num{319} and ReLU as activation function.
For PPO, we set the epochs of updating network to \num{12}, the clipping parameter $\epsilon_{PPO}$ to \num{0.541}, and the coefficient for policy loss and value function to \num{0.496} and \num{0.7918}.
We used a linear scheduler for the learning rate and the entropy coefficient who decay linearly each parameter from \num{6.861e-4} to \num{7.783e-5} and from \num{2.042e-3} to \num{2.458e-4} respectively.
We set the discount factor $\gamma$ to \num{1}, and use the Adam optimizer. We do rollouts of size \num{704}, i.e., more than one episode per iteration, and train with mini-batches of the size \num{33000}.

\section{Experiments}
\label{experiment}

\begin{table*}[t]
\caption{Make-span of solutions found by different approaches on Taillard's and Demirkol's instances. While we do not achieve the same performance as the best tool on the market (OR-Tools), our approach performs better than dispatching rules and previous RL literature.}
\label{solution-makespan}
\begin{center}
\begin{small}
\begin{sc}
\ra{1.2}
\begin{tabular}{llrrrrrrr}
\toprule
Dataset & Instance & Ours & FIFO  & MWKR & \cite{zhang2020learning} & \cite{9218934} & \thead{OR\\Tools} & \thead{Upper\\Bound} \\
\midrule
Taillard & ta41 &\bfseries{2208} &\num{2543} &\num{2632} &\num{2667} &\num{2450} &\num{2144} &\num{2005} \\
& ta42 &\bfseries{2168} &\num{2578} &\num{2401} &\num{2664} &\num{2351} &\num{2071} &\num{1937} \\
& ta43 &\bfseries{2086} &\num{2506} &\num{2385} &\num{2431} &--- &\num{1967} &\num{1846} \\
& ta44 &\bfseries{2261} &\num{2555} &\num{2532} &\num{2714} &--- &\num{2094} &\num{1979} \\
& ta45 &\bfseries{2227} &\num{2565} &\num{2431} &\num{2637} &--- &\num{2032} &\num{2000} \\
& ta46 &\bfseries{2349} &\num{2617} &\num{2485} &\num{2776} &--- &\num{2129} &\num{2004} \\
& ta47 &\bfseries{2101} &\num{2508} &\num{2301} &\num{2476} &--- &\num{1952} &\num{1889} \\
& ta48 &\bfseries{2267} &\num{2541} &\num{2350} &\num{2490} &--- &\num{2091} &\num{1941} \\
& ta49 &\bfseries{2154} &\num{2550} &\num{2474} &\num{2556} &--- &\num{2089} &\num{1961} \\
& ta50 &\bfseries{2216} &\num{2531} &\num{2496} &\num{2628} &--- &\num{2010} &\num{1923} \\
& \bf Average &\bfseries{2203} &\bfseries{2549} &\bfseries{2449} &\bfseries{2604} &\bf --- &\bfseries{2058} &\bfseries{1948} \\
\\
Demirkol & dmu16 &\bfseries{4188} &\num{4934} &\num{4550} &\num{4953} &\num{4414} &\num{3903} &\num{3751} \\
& dmu17 &\bfseries{4274} &\num{5014} &\num{4874} &\num{5379} &--- &\num{3960} &\num{3814} \\
& dmu18 &\bfseries{4326} &\num{4936} &\num{4792} &\num{5100} &--- &\num{4073} &\num{3844} \\
& dmu19 &\bfseries{4195} &\num{4902} &\num{4842} &\num{4889} &--- &\num{3922} &\num{3764} \\
& dmu20 &\bfseries{4074} &\num{4539} &\num{4500} &\num{4859} &--- &\num{3913} &\num{3703} \\
&\bf Average &\bfseries{4211} &\bfseries{4865} &\bfseries{4712} &\bfseries{5036} &\bf --- &\bfseries{3954} &\bfseries{3775} \\
\bottomrule
\end{tabular}
\end{sc}
\end{small}
\end{center}
\end{table*}

This section presents our experimental results against public benchmarks.

\subsection{Benchmark Instances}
\label{instances}

We picked one classical set of benchmark instances provided by \citet{Taillard1993}. These well-studies instances are of different sizes with the upper and lower bound solution provided for each instance.\footnote{\url{http://jobshop.jjvh.nl/index.php}}
We have selected the class of 10 instances with \num{30} jobs and \num{20} machines because they are considered as hard (i.e., harder than bigger instances). The optimal solutions are not yet known. Still, the literature provides us a lower and upper bound for each instance.

To assess that our environment's performance does not overfit Taillard's instance, we also select another set of instances of \citet{DEMIRKOL1998137}. We picked the \num{5} instances with, as Taillard's instances, \num{30} jobs and \num{20} machines where the goal is also to minimize the make-span.
Our approach is not constraint by the size of the problem and can be applied to smaller or bigger instances.

\subsection{Baseline Selection}
\label{baselines}

To ensure our agent is learning complex dispatching strategy, we compare it against two well-known dispatching rules from the literature.
FIFO selects the job that has waited the most, and MWKR the job with the most left-over processing time until completeness.
These dispatching rules usually perform quite well as they act greedy (i.e., do not intentionally create holes) and tend to balance job allocation.
Dispatching rules are deterministic and run only once. Therefore, they give a solution in less than a second.

We have selected the two papers from the literature that evaluate their approaches on the same instances as us. In \cite{zhang2020learning}, all the instances we have selected are benchmarked too, so we have added their results to our benchmark. \citet{9218934} have only selected \num{3} out of \num{15} of the instances we have selected. These results are included and only allow for a partial comparison.

To have a good idea about our approach performance, we have include OR-Tools \cite{ortools} the best tool on the market used to solve JSS problem \cite{cp_solver}. We also give the upper solution bounds found in the literature to indicate the solution's qualities.

\subsection{Results}

We conducted a number of experiments on a server with \num{2} Intel Xeon 6138 CPU and a single Nvidia Titan Xp GPU.\footnote{The code is available at \url{https://github.com/prosysscience/JSS}}
Obtained results are presented in Table~\ref{solution-makespan} which shows all found solutions by each approach for every instance of our two datasets. 
As one can see, our approach performs better than dispatching rules as it found better solutions on each instance. On average, our method finds solutions \si{11\percent} better make-span than the best MWKR dispatching rule on Taillard's instances and \si{12\percent} better on Demirkol's one.

These results are interesting as we have used the same hyper-parameters we have used for Taillard's instances, and we have not used Demirkol's instances during the hyper-parameters search. However, these two instances share some similarities related to the problem size (i.e., they have the same number of operations to plan).
It indicates that our approach generalizes well, as it exhibits similar performances on instances from different datasets.

Even static dispatching rules perform quite well in our environment. This can be explained because they act greedy (i.e., they do not take the No-Op action). The dispatching is done with regard to the currently available jobs at a given time-step, not all the jobs, as it is generally the case with dispatching rules when using the disjunctive graph formulation. They also take advantage of the non-final prioritization rules we have implemented.
Outside of our environment, the best dispatching rules MWKR gives an average solution make-span of \num{3193} \cite{zhang2020learning} on Taillard's instances compared to the \num{2449} we have found.

Although we have not reached the performance of cutting edge OR-Tools CP solver who found on average solutions \si{7\percent} better on Taillard's instances and \si{6\percent} better on Demirkol's one; this work focus only on designing an efficient environment and uses a standard DRL algorithm who can be further enhanced to tackle scheduling problems.

We tried to run our algorithm for more than \num{10} minutes, but PPO was stuck in a local minimum (i.e., the average solution make-span and the best one converge). Even entropy regularization was not able to prevent this phenomenon. 

\subsection{Comparison with Literature Results}

It is often difficult to compare different RL approaches as the goal/settings/algorithm used are different. We will put these results in perspective in this sub-section.

In \cite{zhang2020learning}, authors have not trained their agents on the benchmark instances but on randomly generated ones, since their goal was to obtain an agent that generalizes beyond the presented instances. This is a different goal than ours, which is more traditional for combinatorial problems as it is instance-specific and does not reuse the knowledge gained from other instances. However, we obtain better results with static dispatching rules on our environment than their RL agents applied without training. This observation highlights the quality of our environment.
In fact, our agents found \si{18\percent} and \si{20\percent} better solutions than of \citeauthor{zhang2020learning} on Taillard's or Demirkol's instances, respectively.

The comparison with \cite{9218934} is fairer as the authors also train their agent on the benchmark instances. In all three of our common instances, our approach outperforms theirs. For the two Taillard's instances and the unique Demirkol's instance, we obtain \si{10\percent} and \si{5\percent} better solutions, respectively.
Nevertheless, they used different RL algorithms (Dueling Double DQN, an off-policy, value-based algorithm) with different settings. Their training is limited to \num{8000} episodes and not by a time limit.

\section{Conclusions and Future Work}
In this paper, we have built an optimized environment and present an end-to-end DRL based method to automatically learn to dispatch jobs in a time constraint scenario that models industrial exigence. Our benchmark well confirms our method's superiority to the traditional fixed priority dispatching rules and is state-of-the-art performance with respect to the current limited RL literature.
The solutions we were able to find are close to the ones found by one of the best constraint solvers.

This paper focuses mainly on building an efficient environment. In future work, we plan to improve agent exploration to avoid PPO being stuck in a local optimum and explore different transfer learning approaches to enhance our method's performance further.

\bibliography{main}

\begin{thebibliography}{25}
\providecommand{\natexlab}[1]{#1}
\providecommand{\url}[1]{\texttt{#1}}
\providecommand{\urlprefix}{URL }
\expandafter\ifx\csname urlstyle\endcsname\relax
  \providecommand{\doi}[1]{doi:\discretionary{}{}{}#1}\else
  \providecommand{\doi}{doi:\discretionary{}{}{}\begingroup
  \urlstyle{rm}\Url}\fi

\bibitem[{Abadi et~al.(2015)Abadi, Agarwal, Barham, Brevdo, Chen, Citro,
  Corrado, Davis, Dean, Devin, Ghemawat, Goodfellow, Harp, Irving, Isard, Jia,
  Jozefowicz, Kaiser, Kudlur, Levenberg, Man\'{e}, Monga, Moore, Murray, Olah,
  Schuster, Shlens, Steiner, Sutskever, Talwar, Tucker, Vanhoucke, Vasudevan,
  Vi\'{e}gas, Vinyals, Warden, Wattenberg, Wicke, Yu, and
  Zheng}]{tensorflow2015-whitepaper}
Abadi, M.; Agarwal, A.; Barham, P.; Brevdo, E.; Chen, Z.; Citro, C.; Corrado,
  G.~S.; Davis, A.; Dean, J.; Devin, M.; Ghemawat, S.; Goodfellow, I.; Harp,
  A.; Irving, G.; Isard, M.; Jia, Y.; Jozefowicz, R.; Kaiser, L.; Kudlur, M.;
  Levenberg, J.; Man\'{e}, D.; Monga, R.; Moore, S.; Murray, D.; Olah, C.;
  Schuster, M.; Shlens, J.; Steiner, B.; Sutskever, I.; Talwar, K.; Tucker, P.;
  Vanhoucke, V.; Vasudevan, V.; Vi\'{e}gas, F.; Vinyals, O.; Warden, P.;
  Wattenberg, M.; Wicke, M.; Yu, Y.; and Zheng, X. 2015.
\newblock {TensorFlow}: Large-Scale Machine Learning on Heterogeneous Systems.
\newblock \urlprefix\url{https://www.tensorflow.org/}.
\newblock Software available from tensorflow.org.

\bibitem[{Bengio, Lodi, and Prouvost(2021)}]{BENGIO2021405}
Bengio, Y.; Lodi, A.; and Prouvost, A. 2021.
\newblock Machine learning for combinatorial optimization: A methodological
  tour d’horizon.
\newblock \emph{European Journal of Operational Research} 290(2): 405 -- 421.
\newblock ISSN 0377-2217.
\newblock \doi{https://doi.org/10.1016/j.ejor.2020.07.063}.
\newblock
  \urlprefix\url{http://www.sciencedirect.com/science/article/pii/S0377221720306895}.

\bibitem[{Biewald(2020)}]{wandb}
Biewald, L. 2020.
\newblock Experiment Tracking with Weights and Biases.
\newblock \urlprefix\url{https://www.wandb.com/}.
\newblock Software available from wandb.com.

\bibitem[{Chen and Liu(2018)}]{chen2018lifelong}
Chen, Z.; and Liu, B. 2018.
\newblock Lifelong machine learning.
\newblock \emph{Synthesis Lectures on Artificial Intelligence and Machine
  Learning} 12(3): 24--25.

\bibitem[{Da~Col and Teppan(2019)}]{cp_solver}
Da~Col, G.; and Teppan, E.~C. 2019.
\newblock Industrial Size Job Shop Scheduling Tackled by Present Day CP
  Solvers.
\newblock In Schiex, T.; and de~Givry, S., eds., \emph{Principles and Practice
  of Constraint Programming}, 144--160. Cham: Springer International
  Publishing.
\newblock ISBN 978-3-030-30048-7.

\bibitem[{Demirkol, Mehta, and Uzsoy(1998)}]{DEMIRKOL1998137}
Demirkol, E.; Mehta, S.; and Uzsoy, R. 1998.
\newblock Benchmarks for shop scheduling problems.
\newblock \emph{European Journal of Operational Research} 109(1): 137 -- 141.
\newblock ISSN 0377-2217.
\newblock \doi{https://doi.org/10.1016/S0377-2217(97)00019-2}.
\newblock
  \urlprefix\url{http://www.sciencedirect.com/science/article/pii/S0377221797000192}.

\bibitem[{{Han} and {Yang}(2020)}]{9218934}
{Han}, B.~A.; and {Yang}, J.~J. 2020.
\newblock Research on Adaptive Job Shop Scheduling Problems Based on Dueling
  Double DQN.
\newblock \emph{IEEE Access} 8: 186474--186495.
\newblock \doi{10.1109/ACCESS.2020.3029868}.

\bibitem[{Huang and Ontañón(2020)}]{Huang2020}
Huang, S.; and Ontañón, S. 2020.
\newblock A Closer Look at Invalid Action Masking in Policy Gradient
  Algorithms.
\newblock \emph{Preprint} \urlprefix\url{http://arxiv.org/abs/2006.14171}.

\bibitem[{J.F.~Muth(1966)}]{eco1996}
J.F.~Muth, G.~T. 1966.
\newblock Industrial Scheduling.
\newblock \emph{Louvain Economic Review} 32(2): 121–122.
\newblock \doi{10.1017/S0770451800056682}.

\bibitem[{Liang et~al.(2018)Liang, Liaw, Nishihara, Moritz, Fox, Goldberg,
  Gonzalez, Jordan, and Stoica}]{liang2018rllib}
Liang, E.; Liaw, R.; Nishihara, R.; Moritz, P.; Fox, R.; Goldberg, K.;
  Gonzalez, J.~E.; Jordan, M.~I.; and Stoica, I. 2018.
\newblock {RLlib}: Abstractions for Distributed Reinforcement Learning.
\newblock In \emph{International Conference on Machine Learning ({ICML})}.

\bibitem[{Lin et~al.(2019)Lin, Deng, Chih, and Chiu}]{dqnScheduling}
Lin, C.-C.; Deng, D.-J.; Chih, Y.-L.; and Chiu, H.-T. 2019.
\newblock Smart Manufacturing Scheduling With Edge Computing Using Multiclass
  Deep Q Network.
\newblock \emph{IEEE Transactions on Industrial Informatics} 15: 4276--4284.
\newblock \doi{10.1109/TII.2019.2908210}.

\bibitem[{Liu, Chang, and Tseng(2020)}]{Liu2020}
Liu, C.~L.; Chang, C.~C.; and Tseng, C.~J. 2020.
\newblock Actor-critic deep reinforcement learning for solving job shop
  scheduling problems.
\newblock \emph{IEEE Access} 8: 71752--71762.
\newblock ISSN 21693536.
\newblock \doi{10.1109/ACCESS.2020.2987820}.

\bibitem[{Mnih et~al.(2016)Mnih, Badia, Mirza, Graves, Lillicrap, Harley,
  Silver, and Kavukcuoglu}]{Mnih2016}
Mnih, V.; Badia, A.~P.; Mirza, M.; Graves, A.; Lillicrap, T.~P.; Harley, T.;
  Silver, D.; and Kavukcuoglu, K. 2016.
\newblock Asynchronous Methods for Deep Reinforcement Learning.
\newblock \emph{33rd International Conference on Machine Learning, ICML 2016}
  4: 2850--2869.
\newblock \urlprefix\url{http://arxiv.org/abs/1602.01783}.

\bibitem[{Mnih et~al.(2013)Mnih, Kavukcuoglu, Silver, Graves, Antonoglou,
  Wierstra, and Riedmiller}]{atari}
Mnih, V.; Kavukcuoglu, K.; Silver, D.; Graves, A.; Antonoglou, I.; Wierstra,
  D.; and Riedmiller, M. 2013.
\newblock Playing Atari with Deep Reinforcement Learning.
\newblock \emph{NIPS Deep Learning Workshop 2013}
  \urlprefix\url{http://arxiv.org/abs/1312.5602}.

\bibitem[{OpenAI(2017)}]{PPOOpenAI}
OpenAI. 2017.
\newblock Proximal Policy Optimization — Spinning Up documentation.
\newblock
  \urlprefix\url{https://spinningup.openai.com/en/latest/algorithms/ppo.html}.

\bibitem[{Perron and Furnon(2019)}]{ortools}
Perron, L.; and Furnon, V. 2019.
\newblock OR-Tools.
\newblock \urlprefix\url{https://developers.google.com/optimization/}.

\bibitem[{Schulman et~al.(2015)Schulman, Levine, Moritz, Jordan, and
  Abbeel}]{Schulman2015}
Schulman, J.; Levine, S.; Moritz, P.; Jordan, M.~I.; and Abbeel, P. 2015.
\newblock Trust Region Policy Optimization.
\newblock \emph{32nd International Conference on Machine Learning, ICML 2015}
  3: 1889--1897.
\newblock \urlprefix\url{http://arxiv.org/abs/1502.05477}.

\bibitem[{Schulman et~al.(2017)Schulman, Wolski, Dhariwal, Radford, and
  Klimov}]{Schulman2017Ppo}
Schulman, J.; Wolski, F.; Dhariwal, P.; Radford, A.; and Klimov, O. 2017.
\newblock Proximal Policy Optimization Algorithms.
\newblock \emph{CoRR} abs/1707.06347.
\newblock
  \urlprefix\url{http://dblp.uni-trier.de/db/journals/corr/corr1707.html#SchulmanWDRK17}.

\bibitem[{Silver et~al.(2018)Silver, Hubert, Schrittwieser, Antonoglou, Lai,
  Guez, Lanctot, Sifre, Kumaran, Graepel, Lillicrap, Simonyan, and
  Hassabis}]{Silver2017}
Silver, D.; Hubert, T.; Schrittwieser, J.; Antonoglou, I.; Lai, M.; Guez, A.;
  Lanctot, M.; Sifre, L.; Kumaran, D.; Graepel, T.; Lillicrap, T.; Simonyan,
  K.; and Hassabis, D. 2018.
\newblock A general reinforcement learning algorithm that masters chess, shogi,
  and Go through self-play.
\newblock \emph{Science} 362(6419): 1140--1144.
\newblock \urlprefix\url{http://arxiv.org/abs/1712.01815}.

\bibitem[{Sutton and Barto(2018)}]{sutton2018reinforcement}
Sutton, R.~S.; and Barto, A.~G. 2018.
\newblock \emph{Reinforcement learning: An introduction}, 321--338.
\newblock MIT press.

\bibitem[{Taillard(1993)}]{Taillard1993}
Taillard, E. 1993.
\newblock Benchmarks for basic scheduling problems.
\newblock \emph{European Journal of Operational Research} 64: 278--285.

\bibitem[{Vinyals et~al.(2019)Vinyals, Babuschkin, Czarnecki, Mathieu, Dudzik,
  Chung, Choi, Powell, Ewalds, Georgiev, Oh, Horgan, Kroiss, Danihelka, Huang,
  Sifre, Cai, Agapiou, Jaderberg, Vezhnevets, Leblond, Pohlen, Dalibard,
  Budden, Sulsky, Molloy, Paine, Gulcehre, Wang, Pfaff, Wu, Ring, Yogatama,
  Wünsch, McKinney, Smith, Schaul, Lillicrap, Kavukcuoglu, Hassabis, Apps, and
  Silver}]{Vinyals2019}
Vinyals, O.; Babuschkin, I.; Czarnecki, W.~M.; Mathieu, M.; Dudzik, A.; Chung,
  J.; Choi, D.~H.; Powell, R.; Ewalds, T.; Georgiev, P.; Oh, J.; Horgan, D.;
  Kroiss, M.; Danihelka, I.; Huang, A.; Sifre, L.; Cai, T.; Agapiou, J.~P.;
  Jaderberg, M.; Vezhnevets, A.~S.; Leblond, R.; Pohlen, T.; Dalibard, V.;
  Budden, D.; Sulsky, Y.; Molloy, J.; Paine, T.~L.; Gulcehre, C.; Wang, Z.;
  Pfaff, T.; Wu, Y.; Ring, R.; Yogatama, D.; Wünsch, D.; McKinney, K.; Smith,
  O.; Schaul, T.; Lillicrap, T.; Kavukcuoglu, K.; Hassabis, D.; Apps, C.; and
  Silver, D. 2019.
\newblock Grandmaster level in StarCraft II using multi-agent reinforcement
  learning.
\newblock \emph{Nature} 575: 350--354.
\newblock ISSN 14764687.
\newblock \doi{10.1038/s41586-019-1724-z}.
\newblock \urlprefix\url{https://doi.org/10.1038/s41586-019-1724-z}.

\bibitem[{Waschneck et~al.(2018)Waschneck, Reichstaller, Belzner, Altenmüller,
  Bauernhansl, Knapp, and Kyek}]{Waschneck2018}
Waschneck, B.; Reichstaller, A.; Belzner, L.; Altenmüller, T.; Bauernhansl,
  T.; Knapp, A.; and Kyek, A. 2018.
\newblock Optimization of global production scheduling with deep reinforcement
  learning.
\newblock \emph{Procedia CIRP} 72: 1264--1269.
\newblock ISSN 22128271.
\newblock \doi{10.1016/j.procir.2018.03.212}.

\bibitem[{Williams(1992)}]{Williams1992}
Williams, R.~J. 1992.
\newblock Simple statistical gradient-following algorithms for connectionist
  reinforcement learning.
\newblock \emph{Machine Learning} 8: 229--256.
\newblock ISSN 0885-6125.
\newblock \doi{10.1007/bf00992696}.
\newblock \urlprefix\url{https://link.springer.com/article/10.1007/BF00992696}.

\bibitem[{Zhang et~al.(2020)Zhang, Song, Cao, Zhang, Tan, and
  Xu}]{zhang2020learning}
Zhang, C.; Song, W.; Cao, Z.; Zhang, J.; Tan, P.~S.; and Xu, C. 2020.
\newblock Learning to Dispatch for Job Shop Scheduling via Deep Reinforcement
  Learning.

\end{thebibliography}

\end{document}